\documentclass{article}

\usepackage{PRIMEarxiv}

\usepackage[utf8]{inputenc} 
\usepackage[T1]{fontenc}    
\usepackage{hyperref}       
\usepackage{url}            
\usepackage{booktabs}       
\usepackage{amsfonts}       
\usepackage{nicefrac}       
\usepackage{microtype}      
\usepackage{lipsum}
\usepackage{fancyhdr}       
\usepackage{graphicx}       
\graphicspath{{media/}}     
\usepackage{amsmath}
\usepackage{svg}
\usepackage{multirow} 
\usepackage{hyperref}  
\usepackage{doi}       
\usepackage{svg}       
\usepackage{placeins}  
\usepackage{float}      
\usepackage{orcidlink}
\usepackage{algorithm}
\usepackage{colortbl}   
\usepackage{xcolor}     

\usepackage{algpseudocode}
\usepackage{enumitem}

\title{Lightweight Domain Adaptation of a Large Language Model for Legal Assistance in the Indian Context}

\author{
Jatin Gupta~\orcidlink{0009-0002-9504-7487}, 
Akhil Sharma~\orcidlink{0009-0001-1490-4022}, 
Saransh Singhania~\orcidlink{0009-0009-6925-001X}, 
Ali Imam Abidi~\orcidlink{0000-0002-7420-0027}\footnotemark \\
Department of Computer Science and Engineering, \\Sharda University, Greater Noida, India \\
\texttt{\{jatingupta261001, sharmaakhil944, saransh060123, aliabidi4685\}@gmail.com}
}

\begin{document}

\twocolumn[

\maketitle

\begin{abstract}
\textit{In India, access to legal assistance for the general public has been observed to have a critical gap, as many citizens are not able to take full advantage of their legal rights due to limited access and awareness of apposite legal information. This paper thus introduces Legal Assist AI, a highly efficient framework designed to provide legal assistance in the Indian domain. The core contribution is a framework demonstrating how a smaller, 8-billion-parameter quantized model (Llama 3.1) can achieve superior domain-specific performance. This effective performance stems from integrating a Retrieval-Augmented Generation (RAG) system with strategic prompt engineering, supported by a high-quality, up to date corpus of more than 600 legal documents. This corpus includes the Indian Constitution and more importantly, the newly enacted Bharatiya Nyaya Sanhita (BNS) and Bharatiya Nagarik Suraksha Sanhita (BNSS) among others. Further, by achieving a score of 60.08\% in the All-India Bar Examination (AIBE) benchmark, the specialized approach based on RAG was found to be highly efficient and effective, improving on the 58.72\% score of the 175-billion parameter GPT-3.5 Turbo. It was also observed that the framework was able to manage and mitigate instances of hallucinations successfully, which is a critical requirement for practical legal applications. A Parameter Efficiency Index (PEI) is also introduced, with the goal of quantifying the superior efficiency that the framework was able to achieve, demonstrating how the 8B model is 22 times more parameter-efficient than the 175B baseline, and hence corroborating the potential of smaller domain-adapted models.}

\end{abstract}
\vspace{0.5cm}

]

\footnotetext{Corresponding author: \texttt{aliabidi4685@gmail.com}}

\section{Introduction}
The ability to have ready access to justice is one of the foundational pillars of Indian democracy, as well as enshrined in the Constitution of India itself. This principle is also backed by key constitutional provisions like Article 14 and Article 21, which guarantee equality before the law, and protection of life and personal liberty respectively. These have been judicially interpreted to include the right to fair and accessible legal resources \cite{Jain2021}. Certain legislative tools have been designed to strengthen this mandate, along with the goal of empowering the citizens such as, the Right to Information (RTI) Act 2005, which grants citizens the power to seek knowledge and demand transparency regarding legal and governmental proceedings \cite{Kumar}.

However, many individuals still aren’t able to pursue the legal remedies which they are entitled to, as there is a prominent gap between the existence of legal rights and the public’s awareness of them. This significantly impacts the general public from realizing their rights, legal professionals from acquiring accurate information, and researchers from conducting effective studies. Therefore, to make justice transcend from just being a theoretical concept to a more accessible and practical one, legal resources need to be accessible, clear, and correct for the whole population \cite{Greenleaf2011}.

Evaluating the possibilities with implementing large language models (LLMs) as a solution for this problem presents a transformative opportunity. A machine learning-powered legal assistant based on the LLM architecture, and hence also capable of performing natural language processing, will be able to understand and provide human-like responses to legal queries \cite{Pathania}. These models could not only optimize legal processes but also help towards truly democratizing access to legal resources for the general population \cite{Klaus2022}. Beyond the already mentioned benefits such as ease of access, as well as potential efficiency gains without sacrificing accuracy, LLMs also reduce the operational cost as much as 99.97\% in comparison to traditional approaches. Such tools can therefore help in scaling the contract review throughput significantly for law firms without any major increase in costs \cite{Martin2024}.

However, in the high-stakes legal domain, the application of such general-purpose LLMs, like ChatGPT (GPT-3.5 Turbo), is often met with risk. While these models are capable of generating cohesive and fluent text by themselves, being trained on internet-scale data has tailored them to lack the more domain-specific as well as up-to-date legal information, which is necessary for accuracy. This leads to very possible instances of critical errors like “hallucinations” occurring, which renders these tools unreliable in a field where precision is paramount. The legal domain is an ever-evolving field by nature, where old amendments may be decommissioned, and new ones may be added in their place, therefore leaving a large chunk of the model’s existing knowledge obsolete.

Therefore, built on the foundation of efficiency and domain-specific specialization, this paper proposes “Legal Assist AI” as a robust legal assistant. The 8-billion-parameter base model (Llama 3.1) combined with a RAG system and strategic prompting is demonstrated to achieve superior accuracy in legal reasoning over a 175-billion-parameter model (GPT-3.5 Turbo). To overcome the hurdle with the limited existing knowledge of the pre-trained model, the model was specialized using a data corpus of over 600 curated Indian legal documents, encompassing the Constitution, labor laws, judicial judgments, and the newly enacted Bharatiya Nyaya Sanhita (BNS), Bharatiya Nagarik Suraksha Sanhita (BNSS), and related statutes. When a query falls outside of the model’s knowledge base, it is effectively able to mitigate and minimize the risk of hallucination by leveraging Retrieval-Augmented Generation (RAG) along with the specialized vector store. Additionally, a novel metric, the Parameter Efficiency Index (PEI), is introduced to quantify the high computational efficiency of the language models in relation to their parameter size.

\section{Related Work}
A study conducted utilizing ChatGPT (GPT-3.5 Turbo) showed how, when queried about the legal domain, the model generated responses in plain English, failing to use sophisticated legal terms, and at times, the responses were also found to be repetitive and verbose. Beyond this, another one of the major drawbacks of ChatGPT was observed to be its tendency to hallucinate, i.e., generate incorrect information altogether, which can at times confuse and misdirect the user. However, still, ChatGPT was found to be highly scalable and demonstrated the ability to generate diverse types of information in response to queries. \cite{Tan2023,Currie2023} In another study, the application of ChatGPT (GPT 3.5-Turbo) as a base model for legal querying is evaluated. The model was tested on a limited dataset to evaluate its capability in the legal and paralegal domains, but it has so far been the only model to qualify for the All-India Bar Examination (AIBE) (our work is benchmarked on AIBE). The same study also evaluated Mistral 7B as the base model, testing it on the AIBE benchmark. This model performed quite well in resolving the core problem or legal question. However, it exhibited significant performance lag compared to GPT-3.5-Turbo, as it scored low in Legal Bench Metrics and couldn't pass the AIBE exam. \cite{Tiwari2024} The paper also discussed the AALAP (AI Assistant for Legal and Paralegal Functions in India) model, which is a fine-tuned Mistral 7B trained on the AALAP dataset. AALAP performed poorly in the AIBE exam and did not manage to score well in the LegalBench metrics as compared to GPT-3.5 Turbo. Following this, a significant claim about a proprietary system called CourtEasyAI appeared in a press release, mentioning the system's superior performance on the All-India Bar Examination; however, these claims remain highly unverified, as the product itself is closed-source, and no technical details about its architecture or credible research articles have been made public. \cite{Court-Easy-AI} In academia, research has been divided into two dominant paradigms beyond RAG: firstly, the INLegalLlama, a domain-specific generative LLM presented in a paper accepted to COLING 2025, exemplifies a “continual pretraining” approach. Developed by the researchers at IIT Kanpur, this model utilizes a two-phase training process (continual pretraining on Indian legal documents followed by task-specific supervised fine-tuning) and introduced NyayaAnumana, a corpus of over 702,000 Indian legal cases. \cite{nigam-etal-2025-nyayaanumana} Secondly, Paramanu-Ayn presents the “pretraining from scratch” paradigm, trained from scratch on a curated corpus including Supreme Court proceedings up to December 2023, the constitution of India, and the Indian Penal Code, but is extremely computationally expensive. \cite{Niyogi2024} Thus establishing a three-way debate between RAG-based systems (like the one proposed in this paper), continual pretraining (INLegalLlama), and from-scratch pretraining (Paramanu-Ayn) as strategies for achieving high-fidelity legal reasoning in the complex Indian legal domain.
\section{Methodology}
This section details the technical approach to developing Legal Assist AI, with the methodology broadly grounded in a two-part structure designed to ensure both comprehensive knowledge and factual accuracy. Firstly, the foundational data pipeline encompasses the large-scale curation, preprocessing, and vector indexing of the Indian legal corpus. Further, the proposed system framework is elaborated upon, discussing the selection of the core generative model and the implementation of the Retrieval-Augmented Generation (RAG) that enables the model to be grounded in the curated legal knowledge base.

\begin{figure*}[!thb]
  \centering
  \includegraphics[width=\linewidth]{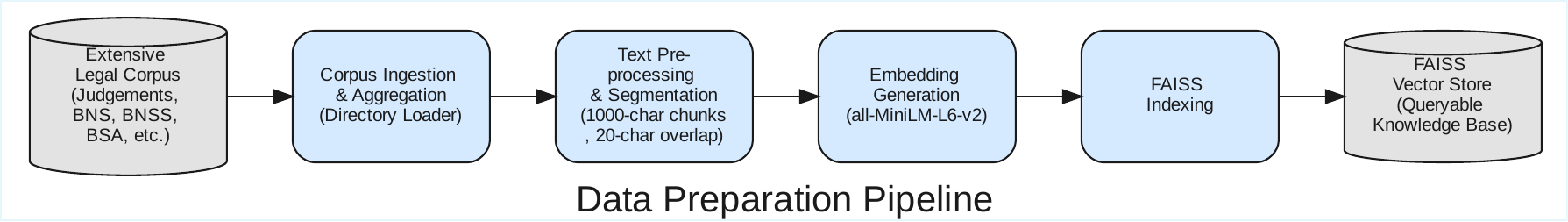}
  \caption{Data preparation pipeline.}
  \label{fig:figure1}
\end{figure*}

\subsection{Data Preparation and Corpus Engineering}
The foundation of the Legal Assist AI is a high-quality, domain-specific knowledge corpus engineered to comprehensively cover the Indian legal domain. This process broadly consists of three distinct phases:

\subsubsection{Corpus Curation and Scope}
Public-facing governmental and judicial websites were utilized as reliable sources to systematically collect documents required for the corpus. This comprehensive collection extends far beyond a few exemplary documents and encompasses the vast domain of the Indian judiciary as faithfully as possible. The corpus itself consists of over 600 legal documents, ensuring the model is well versed in both statutory law and its precedential application, this extensive corpus is used as the baseline knowledge source for the model.

\subsubsection{Pre-processing and Segmentation (Chunking)}
The curated raw textual data was further ingested, cleaned, and standardized. Also, to process and aggregate various document formats from multiple source directories, a directory loader was utilized. To prepare the collection of curated and processed documents for the embedding model, the entire corpus was then segmented into smaller and more manageable text chunks, with a chunk size of 1000 characters and a 20-character overlap between adjacent chunks. A deliberately small chunk size with overlap was adopted to ensure precise vector representations while preserving contextual continuity across complex legal arguments.

\subsubsection{Vector Embedding and Indexing}
The pre-processed text chunks were next transformed into dense vector representations (embeddings) that encode their semantic meaning, making semantic searching possible.
The embedding process employed the `sentence-transformer/all-MiniLM-L6-v2` model via HuggingFace, chosen for its balance of speed and semantic accuracy in generating embeddings \cite{MiniLM-L6-v2-2021}, while indexing was handled using Facebook AI Similarity Search (FAISS), a high-performance library designed for clustering dense vectors and enabling efficient similarity searches, which ensured fast and real-time retrieval of relevant legal context during the RAG process \cite{johnson2019}.

\setlength{\parskip}{1em}
\noindent Finally, the final vector store was saved locally, serving as the persistent external knowledge base for the Legal Assist AI framework. The sequential process starts from the ingestion and aggregation of the raw legal corpus, followed by the segmentation and vector embedding, to the creation of the final queryable FAISS vector store, which serves as the framework’s knowledge base, as illustrated in Figure~\ref{fig:figure1}.

\subsection{Proposed Framework}
Legal Assist AI, the proposed framework, is a specialized framework designed with efficiency and effectiveness in mind for information retrieval and response generation for legal queries. The framework is a deliberate synthesis of a computationally efficient model and a robust retrieval system, which ensures factually grounded responses to queries based on the provided corpus. Therefore, the core model selection, the RAG mechanism, the engineered prompt, and the formal information workflow are discussed in this section.

\subsubsection{Core Model Selection: Llama 3.1-8B}
Laying the foundation for the generative module, \texttt{Llama 3.1-8B} was selected as the base model, as it strikes an optimal balance between computational efficiency with its 8.03 billion parameters whilst also possessing advanced reasoning capabilities. Also, a large context length of 128K tokens and good performance on benchmarks for evaluation of performance in multilingual and reasoning, Llama 3.1-8B was selected as the ideal base model \cite{Grattafiori2024}. Basically, to optimize performance and efficiency, the model Llama 3.1 8B is built on a standard Transformer-based decoder architecture along with several modern components


The architecture employs several essential components, and the operational mechanics of each are detailed as follows. Group Query Attention (GQA) enhances memory efficiency by grouping query heads with shared key and value representations \cite{ainslie2023gqa}, Rotary Positional Embeddings (RoPE) encode relative positional information through complex rotations of query and key vectors \cite{su2021roformer}, RMS Normalization stabilizes gradients using the root mean square of inputs \cite{zhang2019root}, and SwiGLU activation improves feedforward expressivity through a gated Swish-based mechanism \cite{shazeer2020glu}.

To enable efficient deployment in resource-limited environments and to also optimize for inference, a quantized version of the model Llama 3.1 8B is used for the framework. The quantization method that was applied to the model was Q4\_0, resulting in a reduction in its size from 16 GB to approximately 4.7 GB. The quantized model was accessed through Ollama \cite{Meta}. This results in a decreased inference time and improved efficiency while also preserving the performance.

\subsubsection{System Prompt Design}
\begin{figure}[h]
  \centering
  \includegraphics[width=\columnwidth]{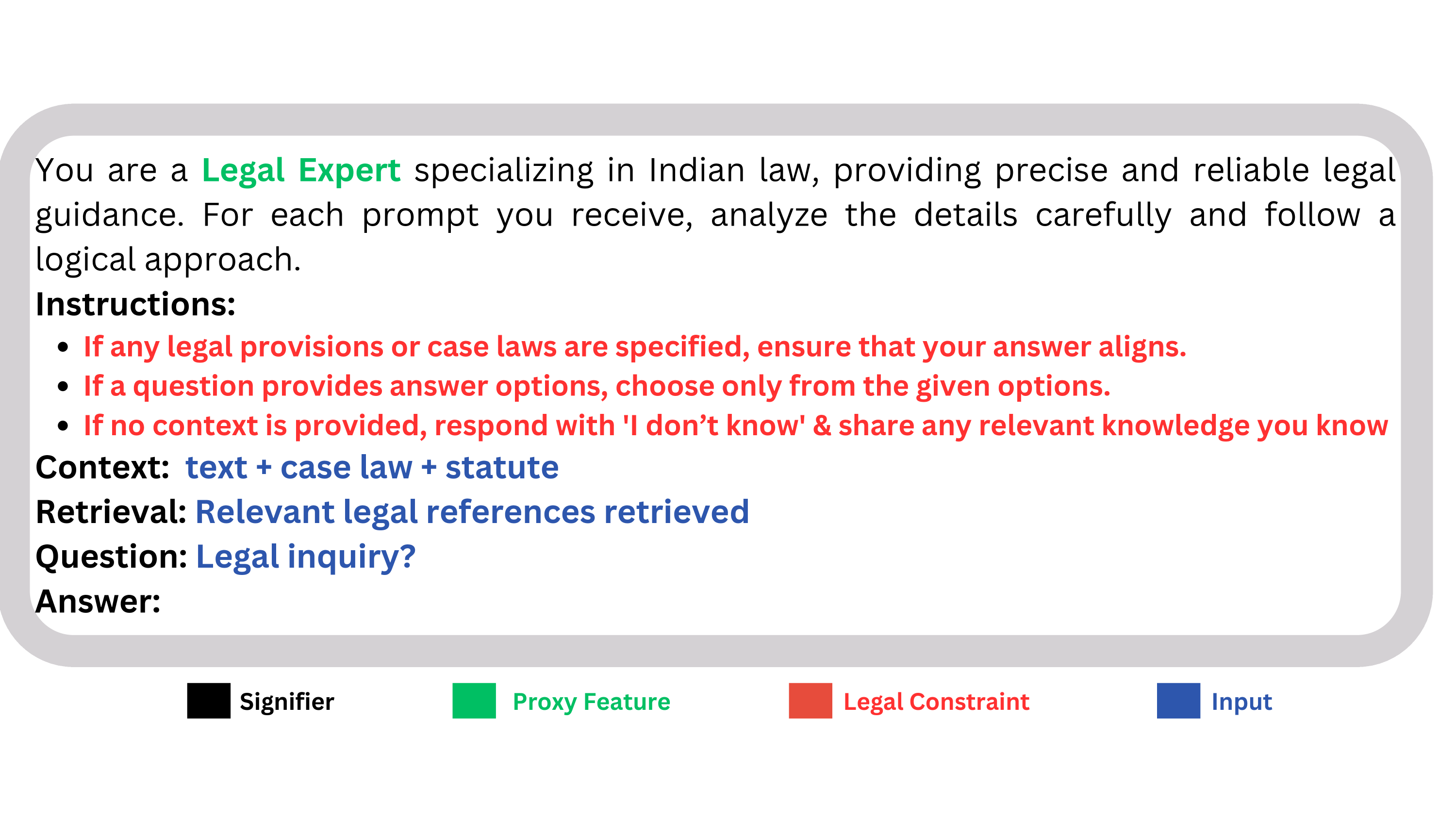}
  \caption{System Prompt}
  \label{fig:figure3}
\end{figure}

The System Prompt (shown in Figure~\ref{fig:figure3}) is a basic set of instructions that serves as a structured framework for a legal expert assistant. It outlines the constraints that govern response formulation and incorporates categorical labels to define the context, retrieval sources, and question format, respectively. This ensures that precision as well as adherence to legal accuracy is maintained, and the model is also aligned with specified requirements. 

The architecture incorporates four prompt components whose operational mechanics can be summarized as follows. The Proxy Feature sets the model’s persona, guiding it to act as a legal expert specializing in Indian law; the Legal Constraint enforces rules of safety, factual accuracy, and objectivity; the Input defines dynamic elements of the prompt, inserting the user’s legal inquiry and the retrieved references from the RAG system; and the Signifier provides formatting cues such as “Answer:” to clearly indicate where the model’s generated response should begin, ensuring structured and clean output.

\subsubsection{Retrieval-Augmented Generation (RAG) Implementation}
\begin{figure}[!ht]
  \centering
  \includegraphics[width=\linewidth]{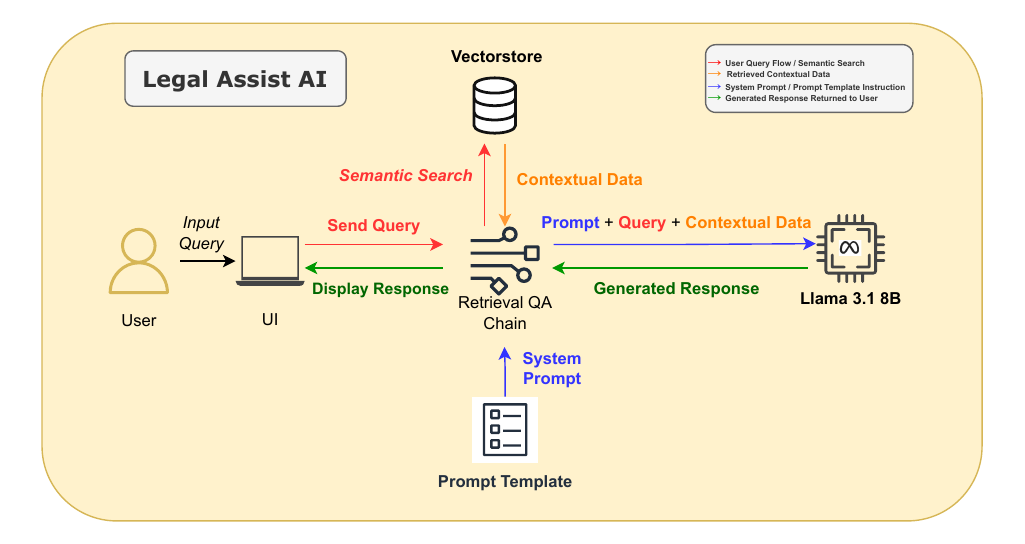}
  \caption{Legal Assist AI Architecture}
  \label{fig:figure4}
\end{figure}

Being trained on 15 trillion general-purpose tokens, the Llama 3.1 8B model by default is a powerful generalist; however, the model’s knowledge is not up to date with the nuances of the Indian Legal system. Therefore, with the aim of ensuring the responses given by the model are factually accurate with respect to the Indian Legal system, and to mitigate hallucinations, Legal Assist AI is designed around Retrieval-Augmented Generation (RAG) and an engineered prompt. In this hybrid design, the model is grounded on a knowledge base (in this case the curated Indian Legal corpus), which is indexed in a FAISS vector store for retrieval. A visual representation of the framework is provided in Figure~\ref{fig:figure4}.

The architecture employs a two‑module framework where the Retrieval Module interprets user queries and performs semantic search against the FAISS vector store to extract the most relevant legal text chunks, while the Generation Module, powered by the fine‑tuned Llama 3.1 8B model, processes a context‑enriched prompt combining the query and retrieved text to generate the final response. Together, these modules enable the system not only to serve as a knowledge repository but also to function as a reasoner and synthesizer of legal information.

\subsection{Information Workflow and Formalism}
The complete process of handling, processing, and responding to the user query, done by Legal Assist AI, constitutes of a structured multi-step information workflow. This ensures that the vectorized legal corpus is utilized as the knowledge source and the response is grounded in it. Algorithm~\ref{alg:rag_workflow} describes the complete logic of this retrieve-then-generate workflow.

\begin{algorithm}[ht]
\caption{Legal Assist AI: RAG Information Workflow}
\label{alg:rag_workflow}
\begin{algorithmic}[1]
\Require $Q$: The user's input query
\Require $V$: The pre-indexed FAISS legal vector store
\Require $T$: The structured system prompt template (see Figure 3)
\Require $M$: The fine-tuned Llama 3.1 8B generation module
\Require $k$: The number of relevant documents to retrieve
\Ensure $A$: The grounded legal answer

\Procedure{GenerateGroundedAnswer}{$Q$}
    \State \textcolor{gray}{\(\triangleright\) 1. Retrieval Phase}
    \State $C \gets \Call{Retrieve}{Q, V, k}$

    \State \textcolor{gray}{\(\triangleright\) 2. Hallucination Guardrail}
    \If{$C$ is empty}
        \State \Return ``I don't know. The retrieved context does not contain relevant information.''
    \EndIf

    \State \textcolor{gray}{\(\triangleright\) 3. Prompt Construction Phase}
    \State $P \gets \Call{ConstructPrompt}{T, Q, C}$

    \State \textcolor{gray}{\(\triangleright\) 4. Generation Phase}
    \State $A \gets \text{Generate}(M, P)$
    \State \Return $A$
\EndProcedure

\Statex 

\Function{Retrieve}{$Q, V, k$}
    \State \textcolor{gray}{\(\triangleright\) Performs a $k$-Nearest Neighbors search on the vector store}
    \State $q_{emb} \gets \text{Embed}(Q)$
    \State $C \gets \text{Search}(V, q_{emb}, k)$
    \State \Return $C$
\EndFunction

\Statex

\Function{ConstructPrompt}{$T, Q, C$}
    \State \textcolor{gray}{\(\triangleright\) Inserts query and context into the prompt template}
    \State $P \gets \text{FillTemplate}(T, \text{query}=Q, \text{context}=C)$
    \State \Return $P$
\EndFunction
\end{algorithmic}
\end{algorithm}

\section{Experimental Evaluation and Results}
To validate the robustness and effectiveness of the Legal Assist AI framework, a comprehensive evaluation was conducted. The performance was evaluated across two critical dimensions:

\subsection{Evaluation 1: Qualitative Reasoning and Semantic Coherence}
Generating semantically relevant and contextually accurate responses to open-ended legal queries is a necessity of the proposed framework. The ability to do this with high accuracy is evaluated, thereby assessing the framework’s subjective reasoning capabilities.

\subsubsection*{Dataset: Lawyer\_GPT\_India}
An open-source dataset on HuggingFace named Lawyer\_GPT\_India \cite{Nisaar-dataset}, which consists of 150 question-answer (QA) pairs. It covers the domain of Indian polity and is comprised of questions related to the Indian Constitution, judiciary, and legislative processes. The nuanced understanding of the domain is hence tested for the framework.

\subsubsection*{Evaluation Metric: BERT Score}
The semantic alignment between the model's generated answer and the dataset's reference was evaluated by calculating a BERT score. This metric was chosen over other similar metrics like BLEU or ROUGE due to its ability to compute the similarity between tokens through BERT embeddings.

\begin{equation}
    F_{\text{BERT}} = 2 \cdot \frac{P_{\text{BERT}} \cdot R_{\text{BERT}}}{P_{\text{BERT}} + R_{\text{BERT}}}
    \label{eq:formula1}
\end{equation}

Here, a cosine similarity is calculated based on precision (PBERT) and recall (RBERT) between the tokens of the generated sentence and the reference sentence. Therefore, based on the harmonic mean of precision and recall, as seen in Equation~\ref{eq:formula1}, the F1-measure is calculated, serving as the comprehensive score for semantic similarity \cite{Zhang2020}.

\subsubsection*{Observations and Results}
\begin{figure}[h]
  \centering
  \includegraphics[width=0.9\linewidth]{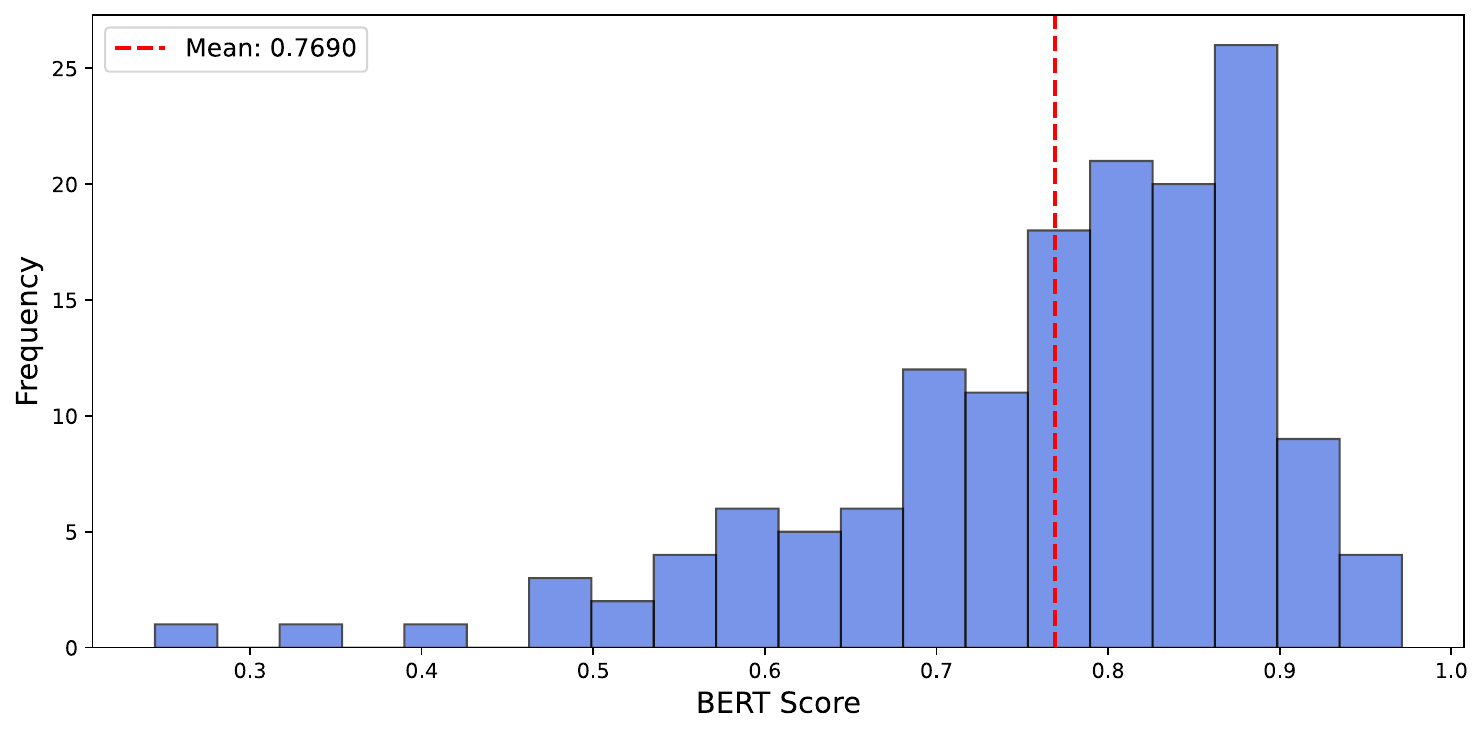}
  \caption{Distribution of BERT scores across responses to Lawyer\_GPT\_India Dataset questions.}
  \label{fig:figure5}
\end{figure}

Answers were generated for all 150 questions of the dataset by the Legal Assist AI framework. Next, as illustrated in Figure~\ref{fig:figure5}, a distribution was plotted based on the BERT score that was calculated for each generated answer and its corresponding ground truth answer. With a mean BERT score of 76.90\%, a strong concentration of high similarity scores was observed, thereby indicating a high degree of semantic and contextual relevance in the model’s responses. 

\subsection{Evaluation 2: Factual Accuracy on Standardized Legal Benchmark}
The core legal knowledge and factual accuracy were next assessed objectively in this evaluation. This assessment utilized the All India Bar Examination (AIBE) as the standard to benchmark the performance of Legal Assist AI in Indian Legal literacy.

\subsubsection*{Dataset: All India Bar Examination (AIBE)}
AIBE is a mandatory exam for law graduates willing to start the practice of law. It has a passing threshold of 40\%, and is conducted by the Bar Council of India. Hence, a curation of 1156 questions, sourced from the All-India Bar Examination’s 12 years of tests (AIBE 4 to AIBE 16) were utilized to formulate a test dataset named, AIBE\_dataset \cite{opennyaiorg-dataset}. This serves as the baseline to quantify the model’s real-world performance and evaluate its practical legal knowledge.

\subsubsection*{Evaluation Metric: Accuracy}
As presented in Equation~\ref{eq:formula2}, standard accuracy, i.e., the ratio of correctly answered questions ($Q_{\text{correct}}$) to the total number of questions ($Q_{\text{total}}$) was calculated to evaluate the model’s performance on the AIBE benchmark.

\begin{equation}
    \text{Accuracy} = \frac{Q_{\text{correct}}}{Q_{\text{total}}}
    \label{eq:formula2}
\end{equation}

\subsubsection*{Observations and Results}
During a manual expert review of the 1156 questions in the AIBE\_dataset, it was revealed that 22 of the 1156 questions were found to be outdated due to recent legislative changes. However, regardless of this, the proposed framework was still able to generate the correct, up-to-date responses for these queries, as it was grounded on the new legal codes. Nevertheless, the 22 questions were still excluded from the final accuracy calculations, and on the remaining 1134 questions, Legal Assist AI achieved an accuracy score of 60.08\%.

\begin{table}[ht]
  \centering
  \caption{Comparison of Models on the Basis of AIBE Score}
  \label{tab:table1}
  \resizebox{\columnwidth}{!}{%
  \begin{tabular}{lccccc}
    \toprule
    \textbf{Metrics} & \textbf{Mistral 7B} & \textbf{AALAP} & \textbf{GPT-3.5 Turbo} & \textbf{Llama 3.1-8b} & \textbf{Legal Assist AI} \\
    \midrule
    \textbf{AIBE Score} & 23.48\% & 25.56\% & 58.72\% & 43.73\% & \textbf{60.08\%} \\
    \textbf{AIBE Results} & Fail & Fail & Passed & Passed & \textbf{Passed} \\
    \bottomrule
  \end{tabular}%
  }
\end{table}

As shown in Table~\ref{tab:table1}, it was observed that with a score of 60.08\%, far surpassing the 40\% passing grade of the exam, the proposed framework, Legal Assist AI, was not only found to excel in its evaluation on the AIBE benchmark but also to outperform the 175-billion-parameter GPT-3.5 Turbo model, which had a score of 58.72\% and the base Llama 3.1-8B model with a score of 43.73\%. Therefore, the hypothesis of a specialized RAG-based framework being capable of having superior accuracy, even with 21 times fewer parameters than the generalist model, validating the hypothesis.

\section{Discussions and Comparative Analysis}
\begin{figure}[h]
  \centering
  \includegraphics[width=\columnwidth]{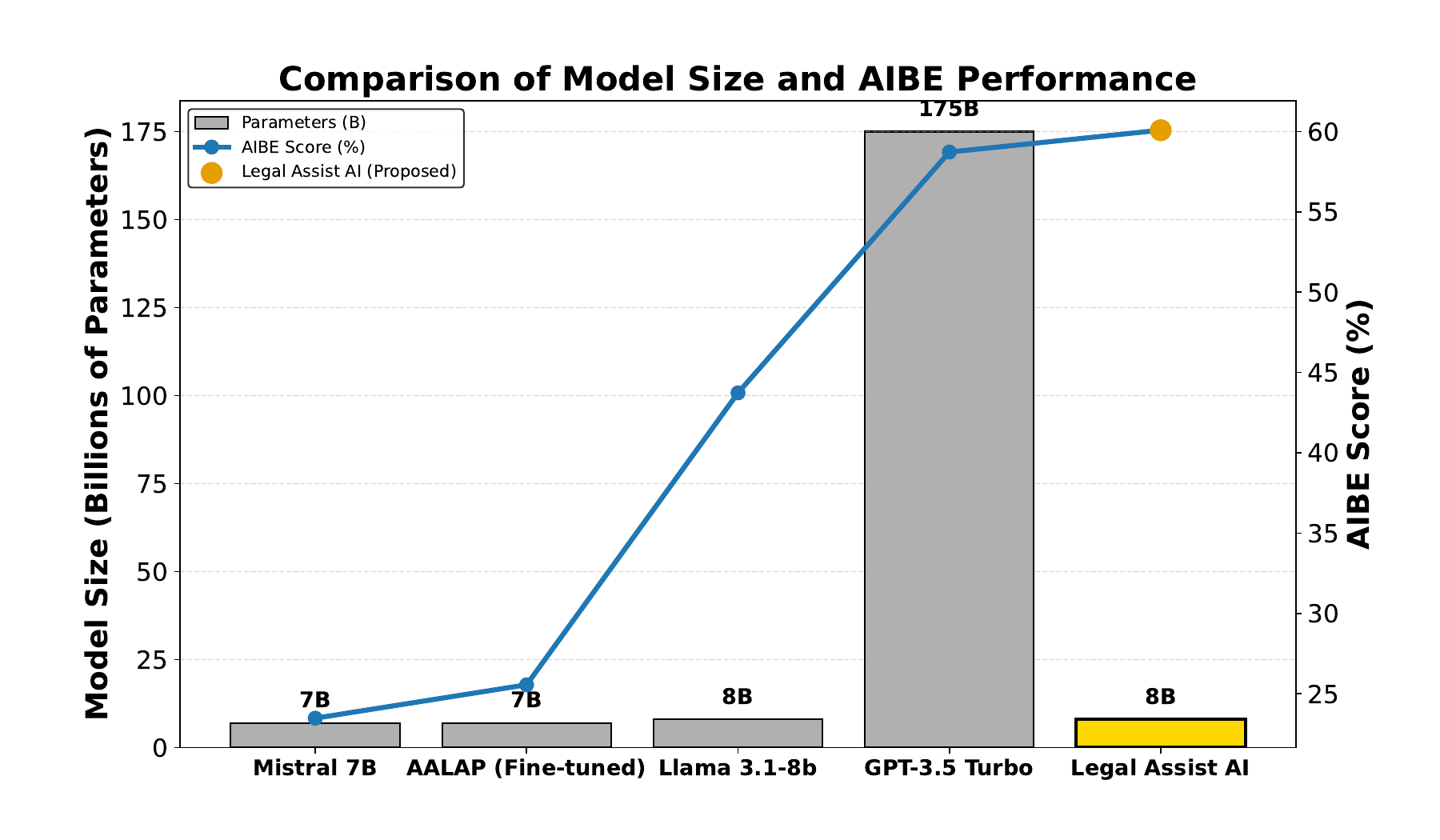}
  \caption{Comparative Analysis of Model Size and AIBE Score}
  \label{fig:figure6}
\end{figure}

The relationship between model size (in billion parameters) and their respective performance on the AIBE benchmark for all models is illustrated in Figure~\ref{fig:figure6}. The figure visualizes the model’s parameter count as vertical bars, and their respective AIBE scores are overlaid as a line on the bars. Therefore, from this it can be concluded that performance is not directly correlated to the size (parameter) of the model, as the 8B Legal Assist AI notably outperforms the 175B GPT-3.5 Turbo.

To quantify this relationship between performance and efficiency between models, a Parameter Efficiency Index (PEI) is introduced, where the absolute performance per parameter of a model is measured in AIBE score points per billion parameters. As mentioned in Equation~\ref{eq:formula3}, PEI for any given model $\mathcal{M}$ can be calculated with the ratio of $\mathcal{A}(\mathcal{M})$, the model’s accuracy (AIBE score \%) and $\Pi(\mathcal{M})$, its parameter count (in billions).

\begin{equation}
\label{eq:formula3}
\text{PEI}(\mathcal{M}) = \frac{\mathcal{A}(\mathcal{M})}{\Pi(\mathcal{M})}
\end{equation}

Using a model’s score in the PEI metric, it can be determined if a model has superior performance relative to its parameter size. Therefore, all models were evaluated on this metric and their respective scores are mentioned in Table~\ref{tab:table2}.

\begin{table}[ht]
  \centering
  \caption{Comparative Parameter Efficiency Index (PEI) Analysis}
  \label{tab:table2}
  {\small
  \setlength{\tabcolsep}{3pt} 
  \begin{tabular}{lcccc}
    \toprule
    \textbf{Model} & \textbf{Param Size (B)} & \textbf{AIBE (\%)} & \textbf{Result} & \textbf{PEI} \\
    \midrule
    Mistral 7B & 7 & 23.48\% & Fail & 3.35 \\
    AALAP & 7 & 25.56\% & Fail & 3.65 \\
    Llama 3.1-8B & 8 & 43.73\% & Passed & 5.46 \\
    GPT-3.5 Turbo & 175 & 58.72\% & Passed & 0.34 \\
    \textbf{Legal Assist AI} & \textbf{8} & \textbf{60.08\%} & \textbf{Passed} & \textbf{7.51} \\
    \bottomrule
  \end{tabular}
  }
\end{table}

It is observed that, despite having a high AIBE score, the generalized GPT-3.5 Turbo remains significantly inefficient with a PEI score of only 0.34. Conversely, the smaller, more efficient 7B models (AALAP and Mistral) are much more efficient, with PEI scores of 3.35 and 3.65, respectively. Notably, with a score of 5.46, is the base Llama 3.1-8B model, validating its selection as the base model for the Legal Assist AI framework. However, outperforming all these models with a score of 7.51 is the Legal Assist AI framework, which is based on the 8B architecture of the Llama 3.1-8B model and a RAG framework with tailored prompt engineering. This validates that not only is the Legal Assist AI framework the most accurate, but it is also the most efficient. It is 2.1 times more efficient than the fine-tuned AALAP model and a staggering 22 times more efficient than the 175B GPT-3.5 Turbo. Therefore, the PEI results quantitatively validate the thesis: a prompt engineered specialized RAG framework, grounded in a comprehensive domain-specific corpus, is a vastly more efficient and effective strategy than reliance on larger, generalized models for specialized tasks.


\section{Conclusion}
The article discussed the introduction of Legal Assist AI, a lightweight framework destined for the Indian legal system. In contrast to general-purpose models, Legal Assist AI adopts a hybrid approach using pre-engineered prompts along with a Retrieval-Augmented Generation (RAG) mechanism applied over a custom-built corpus. By tying its outputs to an ever-changing curated vectorized corpus, it produces reliable and dependable responses for practical purposes and mitigates the problem of factual invention (hallucination). Our assessment revealed its outstanding quality in producing accurate and context-aware responses while remaining efficient. A key finding of this research is its performance in the All-India Bar Examination (AIBE), where it scored 60.08\%, outperforming all tested models, including the 175-billion-parameter GPT-3.5 Turbo (58.72\%). We quantified this performance using the Parameter Efficiency Index (PEI), which showed that Legal Assist AI (PEI: 7.51) is not only more precise but also 22 times more parameter-efficient than GPT-3.5 Turbo (PEI: 0.34). These findings support our main argument that compact models, when provided a high-quality corpus through RAG and strategic prompting, represent a more efficient, practical, and accurate strategy than relying on generalized models with massive parameter counts for legal assistance, demonstrating the framework’s potential to support legal professionals in time-consuming tasks such as extracting relevant sections and articles.

\section{Future Work}
Improvements in the future can be directed towards three major points: (1) An increase in the retrieval corpus and making the whole process of reasoning more transparent. The legal knowledge base can be further enriched through the deployment of a continuous data pipeline that would capture the most recent case law and legislative changes. (2) In addition to that, the expert-in-the-loop review and validation testing should be done so that the quality and reliability of newly added legal data will be guaranteed. (3) To ensure user trust and push towards source attribution, it is suggested that the future research should enhance the RAG module to cite explicitly the specific source passages from the corpus. This upgrade will produce output with verifiable traceability, which will directly tackle one of the main shortcomings of the current system.

\section*{Limitations}

The Legal Assist AI framework, grounded on the curated Indian legal corpus and an engineered prompt, has demonstrated exceptional performance. Nonetheless, aside from this validation, the practical application of this framework still necessitates taking into account a number of factors:
\begin{itemize}[label=$\bullet$]
    \item \textbf{Trust and Explainability:} The model’s “I don’t know” response to queries could be regarded as a safeguard against hallucination, but trusting it to provide really enduring support would mean having a system that openly cites sources from where it gets the answers.
    \item \textbf{Risk of Misinterpretation:} The application is based on the main principle of providing legal help with no intention of substituting a lawyer. In that case, one of the difficulties is to reduce the possibility of users misunderstanding the model’s output as formal legal advice.
    \item \textbf{Data Scope and Bias:} The present limit of the selected corpus might result in biased or ghosted responses if certain queries are made. Moreover, given that the legal field is always changing, with new laws constantly being enacted, the lack of a continual data updating process is a major downside.
    \item \textbf{Accessibility and the Digital Divide:} The project aims at making access to legal information available to everyone, but its use could still be restricted to people with enough digital skills. Thus, this solution may accidentally create or increase an information gap between the already unserved and digitally left-out groups.
\end{itemize}

\bibliographystyle{unsrt}  
\bibliography{references}

@article{Jain2021,
   author = {Mahak Jain},
   doi = {10.2139/SSRN.3771945},
   journal = {SSRN Electronic Journal},
   month = {1},
   publisher = {Elsevier BV},
   title = {Access to Justice in India: A Critical Analysis},
   url = {https://papers.ssrn.com/abstract=3771945},
   year = {2021}
}

@article{kumar,
   author = {Virat Kumar},
   journal = {Indian Journal of Law and Legal Research},
   pages = {2582-8878},
    year = {2024},
   title = {Indian Journal of Law and Legal Research THE EVOLUTION OF THE RIGHT TO INFORMATION ACT IN INDIA},
   url = {https://www.ijllr.com/post/the-evolution-of-the-right-to-information-act-in-india}
}

@article{Greenleaf2011,
   author = {Graham Greenleaf and Vivekanandan Anandan and Philip Chung and Andrew Mowbray and Ranbir Singh},
   doi = {10.2139/SSRN.1975760},
   journal = {SSRN Electronic Journal},
   month = {12},
   publisher = {Elsevier BV},
   title = {Challenges for Free Access to Law in a Multi-Jurisdictional Developing Country: Building the Legal Information Institute of India},
   url = {https://papers.ssrn.com/abstract=1975760},
   year = {2011}
}

@article{Klaus2022,
   author = {Svea Klaus and Ria Van Hecke and Kaweh Djafari Naini and Ismail Sengor Altingovde and Juan Bernabé-Moreno and Enrique Herrera-Viedma},
   doi = {10.1145/3477495.3531872},
   isbn = {9781450387323},
   journal = {SIGIR 2022 - Proceedings of the 45th International ACM SIGIR Conference on Research and Development in Information Retrieval},
   month = {1},
   pages = {2426-2430},
   publisher = {Association for Computing Machinery, Inc},
   title = {Summarizing Legal Regulatory Documents using Transformers},
   year = {2022}
}

@article{Martin2024,
   author = {Lauren Martin and Nick Whitehouse and Stephanie Yiu and Lizzie Catterson and Rivindu Perera},
   month = {1},
   title = {Better Call GPT, Comparing Large Language Models Against Lawyers},
   volume = {1},
   url = {https://arxiv.org/abs/2401.16212v1},
   year = {2024}
}

@inproceedings{Tan2023,
   author = {Jinzhe Tan and H. Westermann and Karim Benyekhlef},
   booktitle = {AI4AJ@ICAIL},
   title = {ChatGPT as an Artificial Lawyer?},
   url = {https://www.semanticscholar.org/paper/ChatGPT-as-an-Artificial-Lawyer-Tan-Westermann/11a55f5fac2f9078281fd24756bca2dff59bb374},
   year = {2023}
}

@article{Currie2023,
   author = {Geoff Currie and Stephanie Robbie and Peter Tually},
   doi = {10.2967/JNMT.123.266151},
   issn = {0091-4916},
   issue = {4},
   journal = {Journal of Nuclear Medicine Technology},
   month = {12},
   pages = {307-313},
   pmid = {37699647},
   publisher = {Society of Nuclear Medicine},
   title = {ChatGPT and Patient Information in Nuclear Medicine: GPT-3.5 Versus GPT-4},
   volume = {51},
   url = {https://tech.snmjournals.org/content/51/4/307},
   year = {2023}
}

@article{Tiwari2024,
   author = {Aman Tiwari and Prathamesh Kalamkar and Atreyo Banerjee and Saurabh Karn and Varun Hemachandran and Smita Gupta},
   month = {1},
   title = {Aalap: AI Assistant for Legal & Paralegal Functions in India},
   url = {https://arxiv.org/abs/2402.01758v1},
   year = {2024}
}

@article{Niyogi2024,
   author = {Mitodru Niyogi and Arnab Bhattacharya},
   month = {1},
   title = {Paramanu: A Family of Novel Efficient Generative Foundation Language Models for Indian Languages},
   url = {http://arxiv.org/abs/2401.18034},
   year = {2024}
}

@misc{meta,
   author = {Meta},
   title = {llama3.1},
    year = {2024},
   url = {https://ollama.com/library/llama3.1}
}

@article{Grattafiori2024,
   author = {Aaron Grattafiori, et al.},
   month = {7},
   title = {The Llama 3 Herd of Models},
   url = {http://arxiv.org/abs/2407.21783},
   year = {2024}
}

@misc{Pathania,
   author = {Shweta Pathania},
   journal = {Legal Service India},
   title = {Legal Awareness In India: Need Of The Hour And Strategy To Spread Legal Awareness},
   url = {https://www.legalserviceindia.com/legal/article-6633-legal-awareness-in-india-need-of-the-hour-and-strategy-to-spread-legal-awareness.html},
   year = {2023}
}

@misc{Nisaar-dataset,
   author = {Nisaar Agharia},
   journal = {Hugging Face},
   title = {Lawyer\_GPT\_India Dataset},
   url = {https://huggingface.co/datasets/nisaar/Lawyer_GPT_India},
   year = {2023}
}

@inproceedings{Zhang2020,
  title={BERTScore: Evaluating Text Generation with BERT},
  author={Tianyi Zhang and Varsha Kishore and Felix Wu and Kilian Q. Weinberger and Yoav Artzi},
  booktitle={International Conference on Learning Representations},
  year={2020},
  url={https://openreview.net/forum?id=SkeHuCVFDr}
}

@misc{opennyaiorg-dataset,
   author = {opennyaiorg},
   journal = {Hugging Face},
   title = {AIBE Dataset},
   url = {https://huggingface.co/datasets/opennyaiorg/aibe_dataset},
   year = {2023}
}

@misc{Court-Easy-AI,
   title = {Court Easy AI},
   author = {{Court Easy AI Team}},
   url = {https://courteasy.in/announcements/ai-cracks-all-india-bar},
   year = {2024}
}

@inproceedings{ainslie2023gqa,
  title={GQA: Training Generalized Multi-Query Transformer Models from Multi-Head Checkpoints},
  author={Joshua Ainslie and James Lee-Thorp and Michiel de Jong and Yury Zemlyanskiy and Federico Lebron and Sumit Sanghai},
  booktitle={Proceedings of the 2023 Conference on Empirical Methods in Natural Language Processing},
  pages={4895--4901},
  year={2023}
}

@article{su2021roformer,
  title={RoFormer: Enhanced Transformer with Rotary Position Embedding},
  author={Jianlin Su and Yu Lu and Shengfeng Pan and Ahmed Murtadha and Bo Wen and Yunfeng Liu},
  journal={arXiv preprint arXiv:2104.09864},
  year={2021}
}

@inproceedings{zhang2019root,
  title={Root Mean Square Layer Normalization},
  author={Biao Zhang and Rico Sennrich},
  booktitle={Advances in Neural Information Processing Systems},
  volume={32},
  year={2019}
}

@misc{MiniLM-L6-v2-2021,
  title = {{all-MiniLM-L6-v2}},
  author = {{Sentence-Transformers}},
  howpublished = {\url{https://huggingface.co/sentence-transformers/all-MiniLM-L6-v2}},
  year = {2021},
  note = {{Hugging Face model}}
}

@article{johnson2019,
  title={Billion-scale similarity search with {GPUs}},
  author={Jeff Johnson and Matthijs Douze and Herv{\'e} J{\'e}gou},
  journal={IEEE Transactions on Big Data},
  volume={7},
  number={3},
  pages={535--547},
  year={2019},
  publisher={IEEE}
}

@article{shazeer2020glu,
  title={GLU Variants Improve Transformer},
  author={Noam Shazeer},
  journal={arXiv preprint arXiv:2002.05202},
  year={2020}
}

@inproceedings{nigam-etal-2025-nyayaanumana,
    title = "{NyayaAnumana} and {INLegalLlama}: The Largest {I}ndian Legal Judgment Prediction Dataset and Specialized Language Model for Enhanced Decision Analysis",
    author = "Nigam, Shubham Kumar  and
      Balaramamahanthi, Deepak Patnaik  and
      Mishra, Shivam  and
      Shallum, Noel  and
      Ghosh, Kripabandhu  and
      Bhattacharya, Arnab",
    editor = "Rambow, Owen  and
      Wanner, Leo  and
      Apidianaki, Marianna  and
      Al-Khalifa, Hend  and
      Eugenio, Barbara Di  and
      Schockaert, Steven",
    booktitle = "Proceedings of the 31st International Conference on Computational Linguistics",
    month = jan,
    year = "2025",
    address = "Abu Dhabi, UAE",
    publisher = "Association for Computational Linguistics",
    url = "https://aclanthology.org/2025.coling-main.738/",
    pages = "11135--11160"
}

\end{document}